# Near-Optimal BRL using Optimistic Local Transitions

**Mauricio Araya-López, Vincent Thomas, Olivier Buffet**  MARAYA|VTHOMAS|BUFFET@LORIA.FR
LORIA, Campus scientifique, BP 239, 54506 Vandœuvre-ls-Nancy CEDEX, FRANCE

## Abstract

Model-based Bayesian Reinforcement Learning (BRL) allows a sound formalization of the problem of acting optimally while facing an unknown environment, i.e., avoiding the exploration-exploitation dilemma. However, algorithms explicitly addressing BRL suffer from such a combinatorial explosion that a large body of work relies on heuristic algorithms. This paper introduces BOLT, a simple and (almost) deterministic heuristic algorithm for BRL which is optimistic about the transition function. We analyze BOLT's sample complexity, and show that under certain parameters, the algorithm is near-optimal in the Bayesian sense with high probability. Then, experimental results highlight the key differences of this method compared to previous work.

## 1. Introduction

Acting in an unknown environment requires trading off *exploration* (acting so as to acquire knowledge) and *exploitation* (acting so as to maximize expected return). *Model-Based Bayesian Reinforcement Learning* (BRL) algorithms achieve this while maintaining and using a probability distribution over possible models (which requires expert knowledge under the form of a prior). These algorithms typically fall within one of the three following classes (Asmuth et al., 2009).

**Belief-lookahead** approaches try to optimally trade off exploration and exploitation by reformulating RL as the problem of solving a POMDP where the state is a pair $\omega = (s, b)$, $s$ being the observed state and $b$ the distribution over the possible models; yet, this problem is intractable, allowing only computationally expensive approximate solutions (Poupart et al., 2006).

**Optimistic** approaches propose exploration mechanisms that explicitly attempt to reduce the model uncertainty (Brafman & Tennenholtz, 2003; Kolter & Ng, 2009; Sorg et al., 2010; Asmuth et al., 2009) by relying on the principle of "optimism in the face of uncertainty".

**Undirected** approaches, such as $\epsilon$-greedy or Boltzmann exploration strategies (Sutton & Barto, 1998), perform exploration actions independent of the current knowledge about the environment.

We focus here on optimistic approaches and, as most research in the field and without loss of generality, we consider uncertainty on the transition function, assuming a known reward function. For some algorithms, recent work proves that they are either PAC-MDP (Strehl et al., 2009)—with high probability they often act as an optimal policy would do (if the MDP model were known)—or PAC-BAMDP (Kolter & Ng, 2009)—with high probability they often act as an ideal belief-lookahead algorithm would do.

This paper first presents background on model-based BRL in Section 2, and on PAC-MDP and PAC-BAMDP analysis in Section 3. Then, Section 4 introduces a novel algorithm, BOLT, which, (1) as BOSS (Asmuth et al., 2009), is optimistic about the transition model—which is intuitively appealing since the uncertainty is about the model—and, (2) as BEB (Kolter & Ng, 2009), is (almost) deterministic—which leads to a better control over this approach. We then prove in Section 5 that BOLT is PAC-BAMDP for infinite horizons, by generalizing previous results known for BEB for finite horizon. Experiments in Section 6 then give some insight as to the practical behavior of these algorithms, showing in particular that BOLT seems less sensitive to parameter tuning than BEB.

## 2. Background

### 2.1. Reinforcement Learning

A *Markov Decision Process* (MDP) (Puterman, 1994) is defined by a tuple $\langle \mathcal{S}, \mathcal{A}, T, R \rangle$ where $\mathcal{S}$ is a finite set of *states*, $\mathcal{A}$ is a finite set of *actions*, the *transition* function $T$ is the probability to transition from state $s$ to state $s'$ when some action $a$ is performed: $T(s, a, s') = Pr(s'|s, a)$, and $R(s, a, s')$ is the instant scalar *reward* obtained during this transition. Reinforcement Learning (RL) (Sutton & Barto, 1998) is the problem of finding an optimal decision policy—a mapping $\pi : \mathcal{S} \mapsto \mathcal{A}$—when the model ($T$ without $R$ in our case) is unknown but while interacting with the system. A typical performance criterion is the expected discounted return

$$V_{\boldsymbol{\mu}}^\pi(s) = E_\pi \left[ \sum_{t=0}^\infty \gamma^t R(s_t, a_t, s_{t+1}) \mid s_0 = s, T = \mu \right],$$

where $\boldsymbol{\mu} \in \mathcal{M}$ is the unknown model and $\gamma \in [0, 1]$ is a discount factor. Under an optimal policy, this state value function verifies the Bellman optimality equation (for all $s \in \mathcal{S}$):

$$V_{\boldsymbol{\mu}}^*(s) = \max_{a \in \mathcal{A}} \sum_{s' \in \mathcal{S}} T(s, a, s') \left[ R(s, a, s') + \gamma V_{\boldsymbol{\mu}}^*(s') \right],$$

and computing this optimal value function allows to derive an optimal policy by behaving in a greedy manner, i.e., by picking actions in $\arg\max_{a \in \mathcal{A}} Q_{\boldsymbol{\mu}}^*(s, a)$, where the state-action value function $Q_{\boldsymbol{\mu}}^*$ is defined as

$$Q_{\boldsymbol{\mu}}^*(s, a) = \sum_{s' \in \mathcal{S}} T(s, a, s') \left[ R(s, a, s') + \gamma V_{\boldsymbol{\mu}}^*(s') \right].$$

Typical RL algorithms either (i) directly estimate the optimal state-action value function $Q_{\boldsymbol{\mu}}^*$ (model-free RL), or (ii) learn $T$ to compute $V_{\boldsymbol{\mu}}^*$ or $Q_{\boldsymbol{\mu}}^*$ (model-based RL). Yet, in both cases, a major difficulty is to pick actions so as to trade off exploitation of the current knowledge and exploration to acquire more knowledge.

### 2.2. Model-based Bayesian RL

We consider here *model-based Bayesian Reinforcement Learning* (Strens, 2000), i.e., model-based RL where the knowledge about the model is represented using a probability distribution $\boldsymbol{b}$ over all possible transition models. An initial prior distribution $\boldsymbol{b}_0 = Pr(\boldsymbol{\mu})$ has to be specified, which is then updated using Bayes rule. At time $t$ the posterior $\boldsymbol{b}_t$ depends on the initial distribution $b_0$ and the state-action history so far $h_t = s_0, a_0, \cdots, s_{t-1}, a_{t-1}, s_t$. This update can be applied sequentially due to the Markov assumption, i.e., at time $t+1$ we only need $\boldsymbol{b}_t$ and the triplet $(s_t, a_t, s_{t+1})$ to compute the new distribution:

$$\boldsymbol{b}_{t+1} = Pr(\boldsymbol{\mu}|h_{t+1}, \boldsymbol{b}_0) = Pr(\boldsymbol{\mu}|s_t, a_t, s_{t+1}, \boldsymbol{b}_t). \quad (1)$$

The distribution $\boldsymbol{b}_t$ is known as the *belief* over the model, and summarizes the information that we have gathered about the model at the current time step.

If we consider the belief as part of the state, the resulting belief-MDP can be solved optimally in theory. Remarkably, modelling RL problems as belief-MDPs provides a sound way of dealing with the exploration-exploitation dilemma, because both objectives are naturally included in the same optimization criterion.

The belief-state can thus be written as $\boldsymbol{\omega} = (s, \boldsymbol{b})$, which defines a Bayes-Adaptive MDP (BAMDP) (Duff, 2002), a special kind of belief-MDP where the belief-state is factored into the (visible) system state and the belief over the (hidden) model. Moreover, due to the integration over all possible models in the value function of the BAMDP, the transition function $T(\omega, a, \omega')$ is given by

$$Pr(\boldsymbol{\omega}'|\boldsymbol{\omega}, a) = Pr(\boldsymbol{b}'|\boldsymbol{b}, s, a, s') E[Pr(s'|s, a)|\boldsymbol{b}],$$

where the first probability is 1 if $\boldsymbol{b}'$ complies with Eq. (1) and 0 else. The optimal Bayesian policy can then be obtained by computing the optimal Bayesian value function (Duff, 2002; Poupart et al., 2006):

$$\begin{aligned}
&\mathbb{V}^*(s, \boldsymbol{b}) \\
&= \max_a \left[ \sum_{s'} E[Pr(s'|s, a)|\boldsymbol{b}](R(s, a, s') + \gamma \mathbb{V}^*(s', \boldsymbol{b}')) \right] \\
&= \max_a \left[ \sum_{s'} T(s, a, s', \boldsymbol{b})(R(s, a, s') + \gamma \mathbb{V}^*(s', \boldsymbol{b}')) \right],
\end{aligned} \quad (2)$$

with $\boldsymbol{b}'$ the posterior after the Bayes update with $(s, a, s')$. For the finite horizon case we can use the same reasoning, so that the optimal value can be computed in theory for a finite or infinite horizon, by performing Bayes updates and computing expectations. However, in practice, computing this value function exactly is intractable due to the large branching factor of the tree expansion.

Here, we are interested in heuristic approaches following the *optimism in the face of uncertainty* principle, which consists in assuming a higher return on the most uncertain transitions. Some of them solve the MDP generated by the expected model (at some stage) with an added exploration reward which favors transitions with lesser known models, as in R-MAX (Brafman & Tennenholtz, 2003), BEB (Kolter & Ng, 2009), or with variance based rewards (Sorg et al., 2010). Another approach, used in BOSS (Asmuth et al., 2009), is to solve, when the model has changed sufficiently, an optimistic estimate of the true MDP (obtained by merging multiple sampled models).

### 2.3. Flat and Structured Priors

The selection of a suitable prior is an important issue in BRL algorithms, because it has a direct impact on the solution quality and computing time. A naive approach is to consider one independent Dirichlet distribution for each state-action transition, known as Flat-Dirichlet-Multinomial prior (FDM), whose pdf is defined as

$$\boldsymbol{b} = f(\boldsymbol{\mu}; \boldsymbol{\theta}) = \prod_{s,a} D(\boldsymbol{\mu}_{s,a}; \boldsymbol{\theta}_{s,a}),$$

where $D(\cdot; \cdot)$ are independent Dirichlet distributions. FDMs can be applied to any discrete state-action MDP, but is only appropriate under the strong assumption of independence of the state-action pairs in the transition function. However, this prior has been broadly used because of its simplicity for computing the Bayesian update and the expected value. Consider that the vector of parameters $\boldsymbol{\theta}$ are the counters of observed transitions, then the expected value of a transition probability is $E[Pr(s'|s,a)|\boldsymbol{b}] = \frac{\boldsymbol{\theta}_{s,a}(s')}{\sum_{s''} \boldsymbol{\theta}_{s,a}(s'')}$, and the Bayesian update under the evidence of a transition $(s, a, s')$, is reduced only to $\boldsymbol{\theta}'_{s,a}(s') = \boldsymbol{\theta}_{s,a}(s') + 1$.

Even though FDMs are useful to analyze and benchmark algorithms, in practice they are inefficient because they do not exploit structured information about the problem. One can for example encode the fact that multiple actions share the same model by factoring multiple Dirichlet distributions, or allow the algorithm to identify such structures using Dirichlet distributions combined using Chinese Restaurant Processes or Indian Buffet Processes (Asmuth et al., 2009).

## 3. PAC Algorithms

*Probably Approximately Correct Learning* (PAC) provides a way of analyzing the quality of learning algorithms (Valiant, 1984). The general idea is that with high probability $1 - \delta$ (probably), a machine with a low training error produces a low generalization error bounded by $\epsilon$ (approximately correct). If the number of steps needed to arrive to this condition is bounded by a polynomial function, then the algorithm is PAC-efficient.

### 3.1. PAC-MDP Analysis

In RL, the PAC-MDP property (Strehl et al., 2009) guarantees that an algorithm generates an $\epsilon$-close policy with probability $1 - \delta$ in all but a polynomial number of steps. An important result is the general PAC-MDP Theorem 10 in Strehl et al. (2009), where three sufficient conditions are presented to comply with the PAC-MDP property. First, the algorithm must use at least near *optimistic* values with high probability. Also, the algorithm must guarantee with high probability that it is *accurate*, meaning that, for the known parts of the model, its actual evaluation will be $\epsilon$-close to the optimal value function. Finally, the number of non-$\epsilon$-close steps (also called *sample complexity*) must be bounded by a *polynomial* function.

In mathematical terms, PAC-MDP algorithms are those for which, with probability $1 - \delta$, the evaluation of a policy $\boldsymbol{A}_t$, generated by algorithm $\boldsymbol{A}$ at time $t$ over the real underlying model $\boldsymbol{\mu}_0$, is $\epsilon$-close to the optimal policy over the same model in all but a polynomial number of steps:

$$V^{\boldsymbol{A}_t}_{\boldsymbol{\mu}_0}(s) \geq V^*_{\boldsymbol{\mu}_0}(s) - \epsilon. \tag{3}$$

Several RL algorithms comply with the PAC-MDP property, differing from one another mainly on the tightness of the sample complexity bound. For example, R-MAX and Delayed Q-Learning (Strehl et al., 2009) are some classic RL algorithms for which this property has been proved, whereas BOSS (Asmuth et al., 2009) is a Bayesian RL algorithm which is also PAC-MDP.

In PAC-MDP analysis the policy produced by an algorithm should be close to the optimal policy derived from the real underlying MDP model. This *utopic* policy (Poupart et al., 2006) cannot be computed, because it is impossible to learn exactly the model with a finite number of samples, but it is possible to reason on the probabilistic error bounds of an approximation to this policy.

### 3.2. PAC-BAMDP Analysis

An alternative to the PAC-MDP approach is to be PAC with respect to the optimal *Bayesian* policy, rather than using the optimal *utopic* policy. We will call this *PAC-BAMDP analysis*, because its aim is to guarantee closeness to the optimal solution of the Bayes-Adaptive MDP. This type of analysis was first introduced in Kolter & Ng (2009), under the name of *near-Bayesian* property, where it is shown that a *modified* version of BEB is PAC-BAMDP for the undiscounted finite horizon case [1].

Let us define how to evaluate a policy in the Bayesian sense:

**Definition 3.1.** *The Bayesian evaluation $\mathbb{V}$ of a policy $\pi$ is the expected value given a distribution over*

---

[1] However, some—rectifiable—errors have been spotted in the proof of near-Bayesianness of BEB in Kolter & Ng (2009), as discussed with the authors.

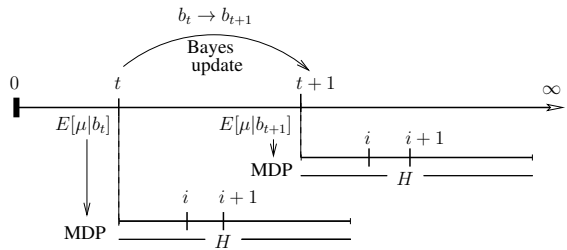

Figure 1. EXPLOIT-like algorithm. At each time step $t$, algorithm performs a Bayes update of the prior, and solves the MDP derived from the expected model of the belief.

models $\boldsymbol{b}$:

$$\mathbb{V}^\pi(s, \boldsymbol{b}) = E_{\boldsymbol{\mu}}[V^\pi_{\boldsymbol{\mu}}(s)|\boldsymbol{b}] = \int_{\mathcal{M}} V^\pi_{\boldsymbol{\mu}}(s) Pr(\boldsymbol{\mu}|\boldsymbol{b}) d\boldsymbol{\mu}.$$

This definition has already been presented implicitly by Duff (2002), but it is very important to point out the difference between a normal MDP evaluation over some known MDP, and the Bayesian evaluation[2]. This definition is consistent with Eq. 2, where

$$\mathbb{V}^*(s, \boldsymbol{b}) = \max_\pi \int_{\mathcal{M}} V^\pi_{\boldsymbol{\mu}}(s) Pr(\boldsymbol{\mu}|\boldsymbol{b}) d\boldsymbol{\mu}$$

$$= \max_a \left[ \sum_{s'} E[Pr(s'|s,a)|\boldsymbol{b}](R(s,a,s') + \gamma \mathbb{V}^*(s', \boldsymbol{b}')) \right].$$

Let us define the PAC-BAMDP property:

**Definition 3.2.** *We say that an algorithm is PAC-BAMDP if, with probability $1 - \delta$, the Bayesian evaluation of a policy $\boldsymbol{A}_t$ generated by algorithm $\boldsymbol{A}$ at time $t$ is $\epsilon$-close to the optimal Bayesian policy in all but a polynomial number of steps, where the Bayesian evaluation is parametrized by the belief $\boldsymbol{b}$:*

$$\mathbb{V}^{\boldsymbol{A}_t}(s, \boldsymbol{b}) \geq \mathbb{V}^*(s, \boldsymbol{b}) - \epsilon, \quad (4)$$

*with $\delta \in [0, 1)$ and $\epsilon > 0$.*

A major conceptual difference is that in PAC-BAMDP analysis, the objective is to guarantee approximate correctness because the optimal Bayesian policy is hard to compute, while in PAC-MDP analysis, the approximate correctness guarantee is needed because the optimal utopic policy is impossible to find in a finite number of steps.

## 4. Optimistic BRL Algorithms

Sec. 2.2 has shown how to theoretically compute the optimal Bayesian value function. This computation being intractable, it is common to use suboptimal—yet efficient—algorithms. A popular technique is to maintain a posterior over the belief, select one representative MDP based on the posterior and act according to its value function. The baseline algorithm in this family is called EXPLOIT (Poupart et al., 2006), where the expected model of $\boldsymbol{b}$ is selected at each time step. Therefore, the algorithm has to solve a different MDP of horizon $H$—an algorithm parameter, not the problem horizon— at each time step $t$ as can be seen in Fig. 1. We will consider for the analysis that $H$ is the number of iterations $i$ that value iteration performs at each time step $t$, but in practice convergence can be reached long before the theoretically derived $H$ for the infinite horizon case.

---

[2]We use a different notation for the Bayesian evaluation, $\mathbb{V}$, to distinguish it from a normal MDP evaluation $V$.

BEB (Kolter & Ng, 2009) follows the same idea as EXPLOIT, but adding an exploration bonus to the reward function. In contrast, BOSS (Asmuth et al., 2009) does not use the EXPLOIT approach, but samples different models from the prior and uses them to construct an optimistic MDP. BEB has the advantage of being an almost deterministic algorithm[3] and does not rely on sampling as BOSS. On the other hand, BOSS is optimistic about the transitions, which is where the uncertainty lies, meanwhile BEB is optimistic about the reward function, even though this function is known.

### 4.1. Bayesian Optimistic Local Transitions

In this section, we introduce a novel algorithm called BOLT (*Bayesian Optimistic Local Transitions*), which relies on acting, at each time step $t$, by following the optimal policy for an optimistic variant of the current expected model. This variant is obtained by, for each state-action pair, optimistically boosting the Bayesian updates before computing the local expected transition model. This is achieved using a new MDP with an augmented action space $\mathfrak{A} = \mathcal{A} \times \mathcal{S}$, where the transition model for action $\alpha = (a, \sigma)$ in state $s$ is the local expected model derived from $\boldsymbol{b}_t$ updated with an artificial evidence of transitions $\lambda^\eta_{s,a,\sigma} = \{(s, a, \sigma), \ldots, (s, a, \sigma)\}$ of size $\eta$ (a parameter of the algorithm). In other words, we pick both an action $a$ plus the next state $\sigma$ we would like to occur with a higher probability. The MDP can be solved as follows:

$$V^{\text{BOLT}}_i(s, \boldsymbol{b}_t)$$
$$= \max_\alpha \sum_{s'} \hat{T}(s, \alpha, s', \boldsymbol{b}_t) \left[ R(s, a, s') + \gamma V^{\text{BOLT}}_{i-1}(s', \boldsymbol{b}_t) \right]$$
$$\text{with} \quad \hat{T}(s, \alpha, s') = E[Pr(s'|s,a)|\boldsymbol{b}_t, \lambda^\eta_{s,a,\sigma}].$$

BOLT's value iteration neglects the evolution of $\boldsymbol{b}_t$, but the modified transition function works as an optimistic

---

[3]In case of equal values, actions are sampled uniformly.

approximation of the neglected Bayesian evolution.

Modifying the transition function seems to be a more natural approach than modifying the reward function as in BEB, since the uncertainty we consider in these problems is about the transition function, not about the reward function.

From a computational point of view, each update in BOLT requires $|\mathcal{S}|$ times more computations than each update in BEB. This implies computation times multiplied by $|\mathcal{S}|$ when solving finite horizon problems using dynamic programming, and probably a similar increase for value iteration. However, under structured priors, not all the next states $\sigma$ must be explored, but only those which are possible transitions.

Here, the optimism is controlled by the positive parameter $\eta$—an integer or real-valued parameter depending on the family of distributions—and the behaviour using different parameter values will depend on the used family of distributions. However, for common priors like FDMs, it can be proved that BOLT is always optimistic with respect to the optimal Bayesian value function.

**Lemma 4.1** (BOLT's Optimism). *Let $(s_t, \boldsymbol{b}_t)$ be the current belief-state from which we apply BOLT's value iteration with an horizon of $H$ and $\eta = H$. Let also $\boldsymbol{b}_t$ be a prior in the FDM family, and let $\mathbb{V}_H(s_t, \boldsymbol{b}_t)$ be the optimal Bayesian value function. Then, we have*

$$V_H^{\text{BOLT}}(s_t, \boldsymbol{b}_t) \geq \mathbb{V}_H(s_t, \boldsymbol{b}_t).$$

[Proof in (Araya-López et al., 2012)]

## 5. Analysis of BOLT

In this section we prove that BOLT is PAC-BAMDP in the discounted infinite horizon case, when using a FDM prior. The other algorithm proved to be PAC-BAMDP is BEB, but the analysis provided in Kolter & Ng (2009) is only for finite horizon domains with an imposed stopping condition for the Bayes update. Therefore, we include in (Araya-López et al., 2012) an analysis of BEB using the results of this section in order to be able to compare these algorithms theoretically afterwards.

By Definition 3.2, we must analyze the policy $\boldsymbol{A}_t$ generated by BOLT at time $t$, i.e., $\boldsymbol{A}_t = \arg\max_\pi V_H^{\text{BOLT},\pi}(s_t)$, and show that, with high probability and for all but a polynomial number of steps, this policy is $\epsilon$-close to the optimal Bayesian policy.

**Theorem 5.1** (BOLT is PAC-BAMDP). *Let $\boldsymbol{A}_t$ denote the policy followed by BOLT at time $t$ with $\eta = H$. Let also $s_t$ and $\boldsymbol{b}_t$ be the corresponding state and belief at that time. Then, with probability at least $1 - \delta$, BOLT is $\epsilon$-close to the optimal Bayesian policy*

$$\mathbb{V}^{\boldsymbol{A}_t}(s_t, \boldsymbol{b}_t) \geq \mathbb{V}^*(s_t, \boldsymbol{b}_t) - \epsilon$$

*for all but $\tilde{O}\left(\frac{|\mathcal{S}||\mathcal{A}|\eta^2}{\epsilon^2(1-\gamma)^2}\right) = \tilde{O}\left(\frac{|\mathcal{S}||\mathcal{A}|H^2}{\epsilon^2(1-\gamma)^2}\right)$ time steps.*
[Proof in Section 5.2]

In the proof we will see that $H$ depends on $\epsilon$ and $\gamma$. Therefore, the sample complexity bound and the optimism parameter $\eta$ will depend only on the desired correctness ($\epsilon,\delta$) and the problem characteristics ($\gamma,|\mathcal{S}|,|\mathcal{A}|$).

### 5.1. Mixed Value Function

To prove that BOLT is PAC-BAMDP we introduce some preliminary concepts and results. First, let us assume for the analysis that we maintain a vector of transition counters $\theta$, even though the priors may be different from FDMs for the specific lemma presented in this section. As the belief is monitored, at each step we can define a set $K = \{(s,a) | \|\boldsymbol{\theta}_{s,a}\| \geq m\}$ of *known state-action pairs* (Kearns & Singh, 1998), i.e., state-action pairs with "enough" evidence. Also, to analyze an EXPLOIT-like algorithm $\boldsymbol{A}$ in general (like EXPLOIT, BOLT or BEB) we introduce a *mixed* value function $\tilde{\mathbb{V}}$ obtained by performing an exact Bayesian update when a state-action pair is in $K$, and $\boldsymbol{A}$'s update when not in $K$. Using these concepts, we can revisit Lemma 5 of Kolter & Ng (2009) for the discounted case.

**Lemma 5.2** (Induced Inequality Revisited). *Let $\mathbb{V}_H^\pi(s_t, \boldsymbol{b}_t)$ be the Bayesian evaluation of a policy $\pi$, $a = \pi(s, \boldsymbol{b})$ be an action from the policy. We define*

$$\tilde{\mathbb{V}}_i^\pi(s, \boldsymbol{b}) = \tag{5}$$
$$\begin{cases} \sum_{s'} T(s,a,s',\boldsymbol{b})(R(s,a,s') + \gamma \tilde{\mathbb{V}}_{i-1}^\pi(s',\boldsymbol{b}')) & \text{if } (s,a) \in K \\ \sum_{s'} \tilde{T}(s,a,s')(\tilde{R}(s,a,s') + \gamma \tilde{\mathbb{V}}_{i-1}^\pi(s',\boldsymbol{b}')) & \text{if } (s,a) \notin K \end{cases}$$

*the mixed value function, where $\tilde{T}$ and $\tilde{R}$ can be different from $T$ and $R$ respectively. Here, $\boldsymbol{b}'$ is the posterior parameter vector after the Bayes update with $(s,a,s')$. Let also $A_K$ be the event that a pair $(s,a) \notin K$ is generated for the first time when starting from state $s_t$ and following the policy $\pi$ for $H$ steps. Assuming normalized rewards for $R$ and a maximum reward $\tilde{R}_{max}$ for $\tilde{R}$, then*

$$\mathbb{V}_H^\pi(s_t, \boldsymbol{b}_t) \geq \tilde{\mathbb{V}}_H^\pi(s_t, \boldsymbol{b}_t) - \frac{(1-\gamma^H)}{(1-\gamma)} \tilde{R}_{max} Pr(A_K),$$
(6)

*where $Pr(A_K)$ is the probability of event $A_K$.*
[Proof in (Araya-López et al., 2012)]

## 5.2. BOLT is PAC-BAMDP

Let $\tilde{\mathbb{V}}_H^{\boldsymbol{A}_t}(s_t, \boldsymbol{b}_t)$ be the evaluation of BOLT's policy $\boldsymbol{A}_t$ using a *mixed* value function where $\tilde{R}(s, a, s') = R(s, a, s')$ the reward function, and $\tilde{T}(s, a, s') = \hat{T}(s, \alpha, s', \boldsymbol{b}_t) = E[Pr(s'|s,a)|\boldsymbol{b}_t, \lambda_{s,a,\sigma}^\eta]$ the BOLT transition model, where $a$ and $\sigma$ are obtained from the policy $\boldsymbol{A}_t$. Note that, even though we apply BOLT's update, we still monitor the belief at each step as presented in Eq. 5. Yet, for $\hat{T}$ we consider the belief at time $t$, and not the monitored belief $\boldsymbol{b}$ as in the Bayesian update

**Lemma 5.3** (BOLT Mixed Bound). *The difference between the optimistic value obtained by BOLT and the Bayesian value obtained by the* mixed *value function under the policy $\boldsymbol{A}_t$ generated by BOLT with $\eta = H$ is bounded by*

$$V_H^{\text{BOLT}}(s_t, \boldsymbol{b}_t) - \tilde{\mathbb{V}}_H^{\boldsymbol{A}_t}(s_t, \boldsymbol{b}_t) \leq \frac{(1-\gamma^H)\eta^2}{(1-\gamma)m}. \quad (7)$$

[Proof in (Araya-López et al., 2012)]

*Proof of Theorem 5.1.* We start by the induced inequality (Lemma 5.2) with $\boldsymbol{A}_t$ the policy generated by BOLT at time $t$, and $\tilde{\mathbb{V}}$ a *mixed* value function using BOLT's update when $(s, a) \notin K$. As $\tilde{R}_{max} = 1$, the chain of inequalities is

$$\mathbb{V}^{\boldsymbol{A}_t}(s_t, \boldsymbol{b}_t) \geq \mathbb{V}_H^{\boldsymbol{A}_t}(s_t, \boldsymbol{b}_t)$$
$$\geq \tilde{\mathbb{V}}_H^{\boldsymbol{A}_t}(s_t, \boldsymbol{b}_t) - \frac{1-\gamma^H}{1-\gamma}Pr(A_K)$$
$$\geq V_H^{\text{BOLT}}(s_t, \boldsymbol{b}_t) - \frac{\eta^2(1-\gamma^H)}{m(1-\gamma)} - \frac{1-\gamma^H}{1-\gamma}Pr(A_K)$$
$$\geq \mathbb{V}_H^*(s_t, \boldsymbol{b}_t) - \frac{\eta^2(1-\gamma^H)}{m(1-\gamma)} - \frac{1-\gamma^H}{1-\gamma}Pr(A_K)$$
$$\geq \mathbb{V}^*(s_t, \boldsymbol{b}_t) - \frac{\eta^2(1-\gamma^H)}{m(1-\gamma)} - \frac{1-\gamma^H}{1-\gamma}Pr(A_K) - \frac{\gamma^H}{(1-\gamma)}$$

where the $3^{rd}$ step is due to Lemma 5.3 (accuracy) and the $4^{th}$ step to Lemma 4.1 (optimism). To simplify the analysis, let us assume that $\frac{\gamma^H}{(1-\gamma)} = \frac{\epsilon}{2}$ and fix $m = \frac{4\eta^2}{\epsilon(1-\gamma)}$.

If $Pr(A_K) > \frac{\eta^2}{m} = \frac{\epsilon(1-\gamma)}{4}$, by the Hoeffding[4] and union bounds we know that $A_K$ occurs no more than

$$O\left(\frac{|\mathcal{S}||\mathcal{A}|m}{Pr(A_K)}\log\frac{|\mathcal{S}||\mathcal{A}|}{\delta}\right) = O\left(\frac{|\mathcal{S}||\mathcal{A}|\eta^2}{\epsilon^2(1-\gamma)^2}\log\frac{|\mathcal{S}||\mathcal{A}|}{\delta}\right)$$

---

[4]Even though the Hoeffding bound assumes that samples are independent, which is trivially not in MDPs, it upper bounds the case where samples are dependent. Recent results shows that tighter bounds can be achieve with a more elaborated analysis (Szita & Szepesvri, 2010).

time steps with probability $1 - \delta$. By neglecting logarithms we have the desired theorem. This bound is derived from the fact that, if $A_K$ occurs more than $|\mathcal{S}||\mathcal{A}|m$ times, then all the state-action pairs are known and we will never escape from $K$ anymore.

For $Pr(A_K) \leq \frac{\eta^2}{m}$, we have that

$$\mathbb{V}^{\boldsymbol{A}_t}(s_t, \boldsymbol{b}_t) \geq \mathbb{V}^*(s_t, \boldsymbol{b}_t) - \frac{\epsilon(1-\gamma^H)}{4} - \frac{\epsilon(1-\gamma^H)}{4} - \frac{\epsilon}{2}$$
$$\geq \mathbb{V}^*(s_t, \boldsymbol{b}_t) - \frac{\epsilon}{4} - \frac{\epsilon}{4} - \frac{\epsilon}{2}$$
$$= \mathbb{V}^*(s_t, \boldsymbol{b}_t) - \epsilon$$

which verifies the proposed theorem. □

Following Kolter & Ng (2009), optimism can be ensured for BEB with $\beta \geq 2H^2$, with $\tilde{O}\left(\frac{|\mathcal{S}||\mathcal{A}|H^4}{\epsilon^2(1-\gamma)^2}\right)$ non $\epsilon$-close steps (see (Araya-López et al., 2012)), which is a worse result than BOLT. Nevertheless, the bounds used in the proofs are loose enough to expect the optimism property to hold for much smaller values of $\beta$ and $\eta$ in practice.

## 6. Experiments

To illustrate the characteristics of BOLT, we present here experimental results over a number of domains. For all the domains we have tried different parameters for BOLT and BEB, but also we have used an $\varepsilon$-greedy variant of EXPLOIT. However, for all the presented problems plain EXPLOIT ($\varepsilon = 0.0$) outperforms the $\varepsilon$-greedy variant.

Please recall that the theoretical values for parameters $\beta$ and $\eta$—that ensure optimism—depend on the horizon $H$ of the MDPs solved at each time step. In these experiments, instead of using this horizon we relied on asynchronous value iteration, stopping when $\|V_{i+1} - V_i\|_\infty < \epsilon$. For solving these infinite MDPs we used $\gamma = 0.95$ and $\epsilon = 0.01$, but be aware that the performance criterion used here is averaged *undiscounted* total rewards.

### 6.1. The Chain Problem

In the 5-state chain problem (Strens, 2000; Poupart et al., 2006), every state is connected to state $s_1$ by taking action $b$ and every state $s_i$ is connected to the next state $s_{i+1}$ with action $a$, except $s_5$ that is connected to itself. At each step, the agent may "slip" with probability $p$, performing the opposite action as intended. Staying in $s_5$ had a reward of 1.0 while coming back to $s_1$ had a reward of 0.2. All other rewards are 0. The priors used for these problems were **Full** (FDM), **Tied**, where the probability $p$ is factored for all transitions, and **Semi**, where each action has an independent factored probability.

| Algorithm | Tied | Semi | Full |
|---|---|---|---|
| EXPLOIT ($\varepsilon = 0$) | 366.1 | 354.9 | 230.2 |
| BEB ($\beta = 1$) | 365.9 | 362.5 | 343.0 |
| BEB ($\beta = 150$) | 366.5 | 297.5 | 165.2 |
| BOLT ($\eta = 7$) | 367.9 | 367.0 | 289.6 |
| BOLT ($\eta = 150$) | 366.6 | 358.3 | 278.7 |
| BEETLE * | 365.0 | 364.8 | 175.4 |
| BOSS * | 365.7 | 365.1 | 300.3 |

Table 1. **Chain Problem results for different priors**. Averaged total reward over 500 trials for an horizon of 1000 with $p = 0.2$. The results with * come from previous publications.

Table 1 shows that BEB outperforms other algorithms with a tuned up $\beta$ value for the FDM prior as already shown by Kolter & Ng (2009). However, for a large value of $\beta$, this performance decreases dramatically. BOLT on the other hand produces results comparable with BOSS for a tuned parameter, but does not decrease too much for a large value of $\eta$. Indeed, this value corresponds to the theoretical bound that ensures optimism, $\eta = H \approx \log(\epsilon(1-\gamma))/\log(\gamma) \approx 150$. Unsurprisingly, the results of BEB and BOLT with informative priors are not much different than other techniques, because the problem degenerates into a easily solvable problem. Nevertheless, BOLT achieves good results for a large $\eta$, in contrast to BEB that fails to provide a competitive result for the Semi prior with large $\beta$.

This variability in the results depending on the parameters, rises the question of the sensitivity to parameter tuning. In a RL domain, one usually cannot tune the algorithm parameters for each problem, because the whole model of the problem is unknown. Therefore, a good RL algorithm must perform well for different problems without modifying its parameters.

Fig. 2 shows how BEB and BOLT behave for different parameters using a **Full** prior. In the low resolution analysis BEB's performance decays very fast, while BOLT also tends to decrease, but maintaining good results. We have also conducted experiments for other values of the slip probability $p$, the same pattern being amplified when $p$ is near 0, i.e., worse decay for BEB and almost constant BOLT results, and obtaining almost identical behavior when $p$ is near 0.5. In the high resolution results BEB goes up and down near 1, while BOLT maintains a similar behaviour as in the low resolution experiment.

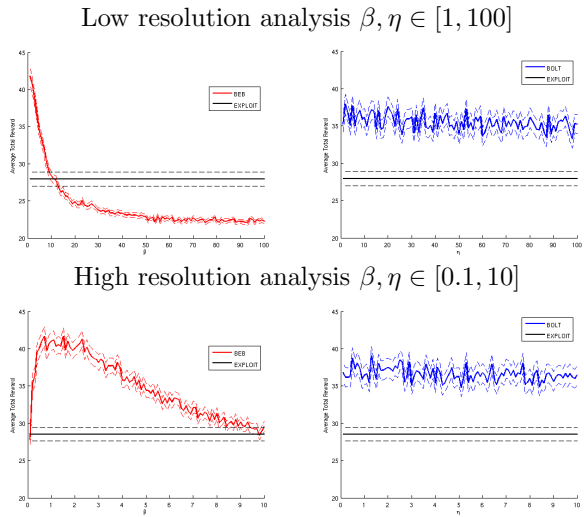

Figure 2. **Chain Problem**. Averaged total reward over 300 trials for an horizon of 150, and for $\beta$ and $\eta$ parameters between 1 and 100, and between 0.1 and 10. As a reference, the value obtained by EXPLOIT is also plotted. All results are shown with a 95% confidence interval.

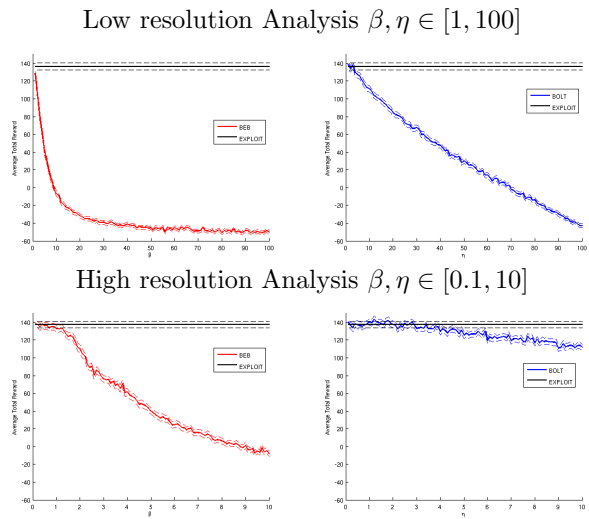

Figure 3. **Paint/Polish Problem**. Averaged total reward over 300 trials for an horizon of 150, for several values of $\beta$ and $\eta$ using an structured prior. As a reference, the value obtained by EXPLOIT is also plotted. All results are shown with a 95% confidence interval.

### 6.2. Other Structured Problems

An other illustrative example is the Paint/Polish problem where the objective is to deliver several polished and painted objects without a scratch, using several stochastic actions with unknown probabilities. The full description of the problem can be found in Walsh et al. (2009). Here, the possible outcomes of each action are given to the agent, but the probabilities of each outcome are not. We have used a structured prior that encodes this information and the results are summarized in Fig. 3, using both high and low resolution analyses. We have also performed this experiment with an FDM prior, obtaining similar results as for the Chain problem. Unsurprisingly, using a structured prior provides better results than using FDMs. However, the high impact of being overoptimistic shown in Fig. 3, does not apply to FDMs, mainly because the learning phase is much shorter using a structured prior. Again, the decay of BEB is much stronger than BOLT, but in contrast to the Chain problem, the best parameter of BOLT beats the best parameter of BEB.

The last example is the Marble Maze problem[5] (Asmuth et al., 2009) where we have explicitly encoded the 16 possible clusters in the prior, leading to little exploration requirements. EXPLOIT provides very good solutions for this problem, and BOLT provides similar results with several different parameters. In contrast, for all the tested $\beta$ parameters, BEB behaves much worse than EXPLOIT. For example, for the best $\eta = 2.0$ BOLT scores $-0.445$, while for the best $\beta = 0.9$ BEB scores $-2.127$, while EXPLOIT scores $-0.590$.

In summary, it is hard to know a priori which algorithm will perform better for a specific problem with a specific prior and given certain parameters. However, BOLT generalizes well (in theory and in practice) for a larger set of parameters, mainly because the optimism is bounded by the probability laws and not by a free parameter as in BEB.

## 7. Conclusion

We have presented BOLT, a novel and simple algorithm that uses an optimistic boost to the Bayes update, which is thus optimistic *about* the uncertainty rather than just *in the face* of uncertainty. We showed that BOLT is strictly optimistic for certain $\eta$ parameters, and used this result to prove that it is also PAC-BAMDP. The sample complexity bounds for BOLT are tighter than for BEB. Experiments show that BOLT is more efficient than BEB when using the theoretically derived parameters in the Chain problem, and in general that BOLT seems more robust to parameter tuning. Future work includes using a dynamic $\eta$ bonus for BOLT, what should be particularly appropriate with finite horizons, and exploring general proofs to guarantee the PAC-BAMDP property for a broader family of priors than FDMs.

---

[5]Averaged over 100 trials with $H = 100$.

## References

Araya-López, M., Thomas, V., and Buffet, O. Near-optimal BRL using optimistic local transitions (extended version). Technical Report 7965, INRIA, May 2012.

Asmuth, J., Li, L., Littman, M.L., Nouri, A., and Wingate, D. A Bayesian sampling approach to exploration in reinforcement learning. In *Proc. of UAI*, 2009.

Brafman, R.I. and Tennenholtz, M. R-max - a general polynomial time algorithm for near-optimal reinforcement learning. *JMLR*, 3:213–231, 2003.

Duff, M. *Optimal learning: Computational procedures for Bayes-adaptive Markov decision processes*. PhD thesis, University of Massachusetts Amherst, 2002.

Kearns, M. and Singh, S. Near-optimal reinforcement learning in polynomial time. In *Machine Learning*, pp. 260–268, 1998.

Kolter, J. and Ng, A. Near-Bayesian exploration in polynomial time. In *Proc. of ICML*, 2009.

Poupart, P., Vlassis, N., Hoey, J., and Regan, K. An analytic solution to discrete Bayesian reinforcement learning. In *Proc. of ICML*, 2006.

Puterman, M. *Markov Decision Processes: Discrete Stochastic Dynamic Programming*. Wiley-Interscience, 1994.

Sorg, J., Singh, S., and Lewis, R. Variance-based rewards for approximate Bayesian reinforcement learning. In *Proc. of UAI*, 2010.

Strehl, A.L., Li, L., and Littman, M.L. Reinforcement learning in finite MDPs: PAC analysis. *JMLR*, 10: 2413–2444, December 2009.

Strens, Malcolm J. A. A Bayesian framework for reinforcement learning. In *Proc. of ICML*, 2000.

Sutton, R. and Barto, A. *Reinforcement Learning: An Introduction*. MIT Press, 1998.

Szita, Istvn and Szepesvri, Csaba. Model-based reinforcement learning with nearly tight exploration complexity bounds. In *Proc. of ICML*, 2010.

Valiant, L. G. A theory of the learnable. In *Proc. of STOC*. ACM, 1984.

Walsh, T.J., Szita, I., Diuk, C., and Littman, M.L. Exploring compact reinforcement-learning representations with linear regression. In *Proc. of UAI*, 2009.